\def\paperlanguage{}
\pdfoutput=1 

\documentclass[letterpaper, 10 pt, conference]{ieeeconf}  

\usepackage{bm}
\usepackage{cite}
\include{preamble}

\IEEEoverridecommandlockouts                              

\title{\LARGE \textbf
  {
    \switchlanguage%
    {%
      Adaptive Whole-body Robotic Tool-use Learning on Low-rigidity Plastic-made Humanoids Using Vision and Tactile Sensors
    }%
    {%
      低剛性樹脂製ヒューマノイドによる視覚・接触覚を利用した\\全身道具利用学習
    }%
  }
}

\author{Kento Kawaharazuka$^{1}$, Kei Okada$^{1}$, and Masayuki Inaba$^{1}$
  \thanks{$^{1}$ The authors are with the Department of Mechano-Informatics, Graduate School of Information Science and Technology, The University of Tokyo, 7-3-1 Hongo, Bunkyo-ku, Tokyo, 113-8656, Japan.
    {\texttt\small [kawaharazuka, k-okada, inaba]@jsk.t.u-tokyo.ac.jp}
  }
}
\begin{document}

\maketitle
\thispagestyle{empty}
\pagestyle{empty}

\begin{abstract}
  \switchlanguage%
  {%
    Various robots have been developed so far; however, we face challenges in modeling the low-rigidity bodies of some robots.
    In particular, the deflection of the body changes during tool-use due to object grasping, resulting in significant shifts in the tool-tip position and the body's center of gravity.
    Moreover, this deflection varies depending on the weight and length of the tool, making these models exceptionally complex.
    However, there is currently no control or learning method that takes all of these effects into account.
    In this study, we propose a method for constructing a neural network that describes the mutual relationship among joint angle, visual information, and tactile information from the feet.
    We aim to train this network using the actual robot data and utilize it for tool-tip control.
    Additionally, we employ Parametric Bias to capture changes in this mutual relationship caused by variations in the weight and length of tools, enabling us to understand the characteristics of the grasped tool from the current sensor information.
    We apply this approach to the whole-body tool-use on KXR, a low-rigidity plastic-made humanoid robot, to validate its effectiveness.
  }%
  {%
    これまで様々なロボットが開発されてきたが, ロボットによっては低剛性で身体のモデリングが困難な場合が多数存在する.
    特に, 道具利用においては物体の把持によって身体のたわみが変化し, 道具の先端位置や身体重心位置等が大きく変化してしまう.
    また, そのたわみ量は道具の重量や長さ等によっても変化するため, これらのモデリングは困難を極める.
    一方で, これらの影響を全て考慮した制御手法, 学習手法は存在しない.
    そこで本研究では, 関節角度や視覚情報, 足裏の接触覚情報等の相互関係を記述するニューラルネットワークを構築し, これらを実機データから学習, 制御に利用する手法を開発する.
    また, 道具の長さや重さの変化によるそれら相互関係の変化をParametric Biasにより捉え, 現在状態から把持した道具の特性を理解することも行う.
    本手法を低剛性な樹脂製ヒューマノイドであるKXRの全身道具利用動作に適用し, その有効性を確認する.
  }%
\end{abstract}

\section{INTRODUCTION}\label{sec:introduction}
\switchlanguage%
{%
  Various types of robots have been developed over the years.
  While there are robots with high rigidity and strong force capabilities, commonly found in industrial applications \cite{hirai1998asimo, kaneko2004hrp2}, there exist low-rigidity robots constructed with rubber or plastic-made links and joints \cite{allgeuer2015igus, makabe2021kxr}.
  The development of these robots can be driven by different factors, such as the pursuit of flexibility in soft robotics \cite{lee2017softrobotics}, cost considerations, or the need for lightweight design.
  While high-rigidity robots are easier to model and can execute precise movements, low-rigidity robots pose challenges in modeling due to the flexibility of their joints and links, which can deform in response to posture and external forces.
  Particularly, although it is possible to apply standard kinematics and dynamics when only the joints have low rigidity, when both joints and links exhibit low rigidity, modeling becomes exceedingly difficult and the application of conventional methods becomes problematic.
  Methods considering the flexibility of these joints and links have been developed \cite{kiang2015flexible, sum2017flexible}, but they still face numerous challenges including the necessity for multiple assumptions and computational complexities when dealing with multi-link systems.
  In tool-use on low-rigidity robots, the deflection of the body changes due to object grasping, leading to significant variations in the tool-tip position and center of gravity.
  Furthermore, the amount of the deflection can vary based on the weight and length of tools, further complicating their modeling.
  In the context of soft robotics, various methods have been developed to learn kinematics and dynamics from flexible and diverse sensors \cite{loo2022multimodal, runge2017fem, thuruthel2019soft}.
  However, there is currently no method capable of handling changes in the body due to variations in tool grasping.
}%
{%
  これまで様々なロボットが開発されてきたが, 産業用ロボットに多く見られる高剛性で力の強いロボットが多数存在する一方\cite{hirai1998asimo, kaneko2004hrp2}, ゴムや樹脂でリンクや関節が構成された低剛性なロボットも存在する\cite{allgeuer2015igus, makabe2021kxr}.
  これらはSoft Robotics \cite{lee2017softrobotics}のように柔軟性を追求して開発される場合もあれば, 値段や軽量化の観点から樹脂によってロボットを構成する場合もあり, その理由はまちまちである.
  高剛性なロボットはモデル化によって容易にかつ正確に動作するが, 低剛性なロボットはその低剛性ゆえに関節やリンクが姿勢や力に応じてたわみ, そのモデリングが難しいという問題がある.
  特に, 関節のみ低剛性な場合は, リンクに関する幾何モデルは正しいとして一般的な運動学や動力学が適用可能であるが, リンクまで低剛性な場合は, よりモデル化が困難を極め, それらの適用が難しくなる.
  もちろん, これら関節やリンクのたわみを考慮したような手法は開発されているが\cite{kiang2015flexible, sum2017flexible}, 多数の仮定が必要な点や, 多リンクにした際の計算量等の観点からまだ問題は多く残る.
  特に, 低剛性なロボットによる道具利用においては, 物体の把持によって身体のたわみ方が変化し, ある姿勢に対する道具の先端位置や重心位置等が大きく変化してしまう.
  また, そのたわみ方は道具の重量や長さ等によっても変化するため, これらのモデリングは非常に難しい.
  Soft Robotの文脈では柔軟で多様なセンサからkinematicsやdynamicsを学習する手法も開発されているが\cite{loo2022multimodal, runge2017fem, thuruthel2019soft}, 道具による身体変化まで扱えている手法はない.
}%

\begin{figure}[t]
  \centering
  \includegraphics[width=1.0\columnwidth]{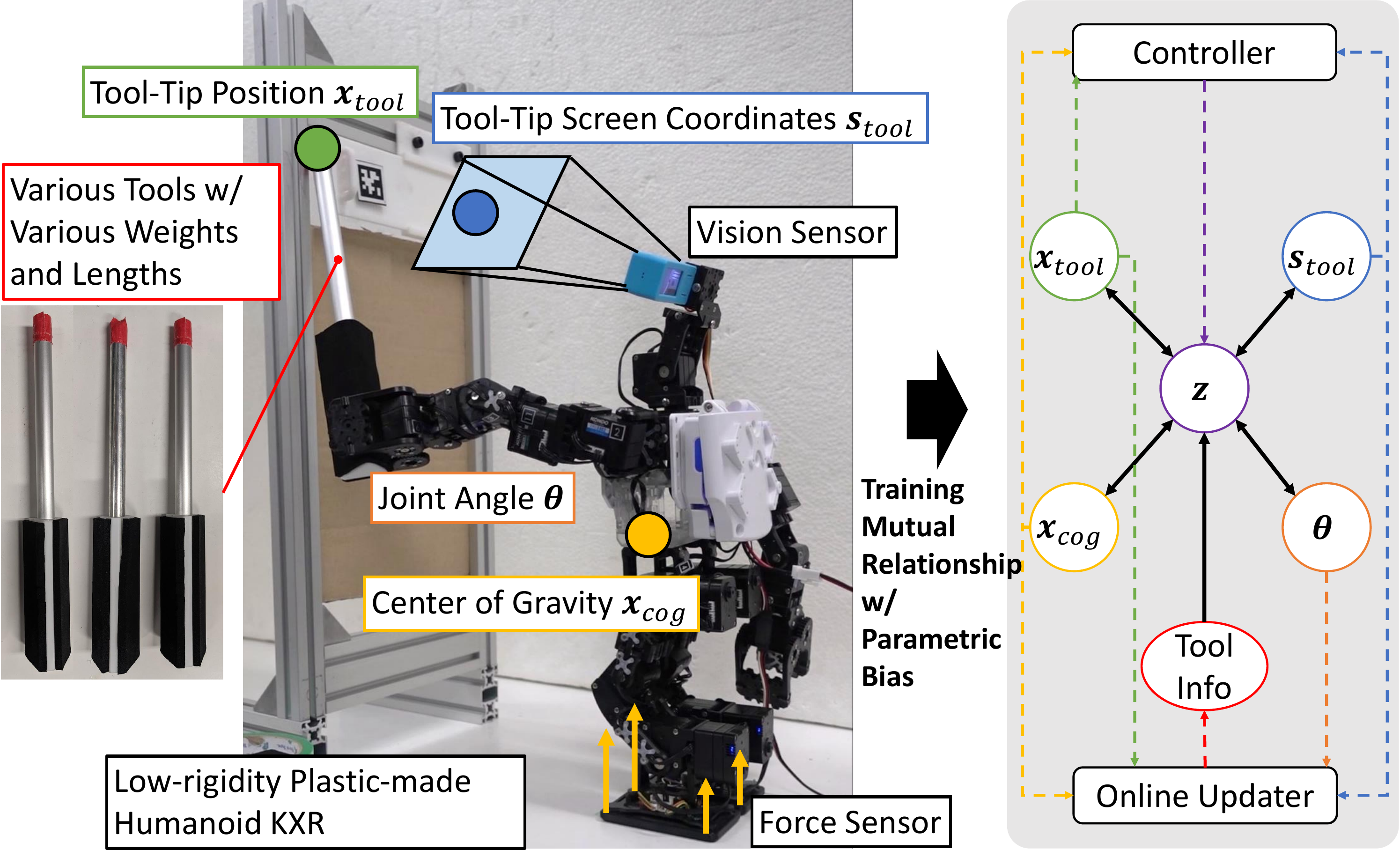}
  \vspace{-3.0ex}
  \caption{The concept of this study: learning the mutual relationship among joint angle, center of gravity, tool-tip position, and tool-tip screen coordinates for adaptive whole-body tool-use of low-rigidity robots considering the changes in tool weight and length.}
  \label{figure:concept}
  \vspace{-3.0ex}
\end{figure}

\begin{figure*}[t]
  \centering
  \includegraphics[width=1.95\columnwidth]{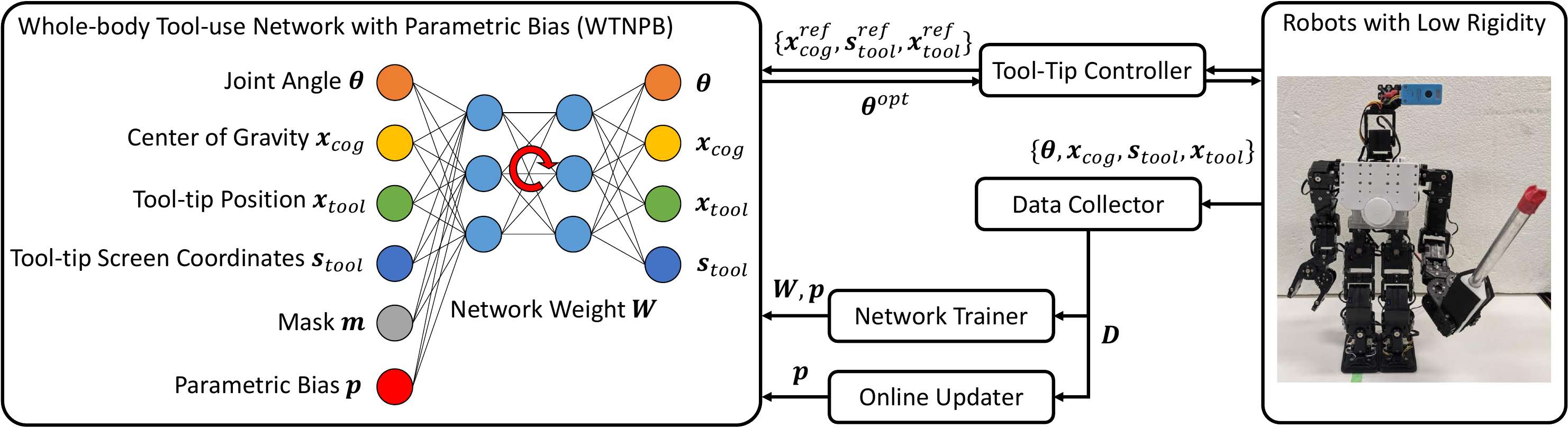}
  \caption{System overview of Whole-body Tool-use Network with Parametric Bias (WTNPB) including Data Collector, Network Trainer, Online Updater, and Tool-Tip Controller for Low-Rigidity Robots.}
  \label{figure:system-overview}
  \vspace{-3.0ex}
\end{figure*}

\switchlanguage%
{%
  In this study, we develop a method for controlling tool-tip position considering deflection, using a neural network that describes the mutual relationship among joint angle, visual information, and tactile information for tool-use on low-rigidity robots (\figref{figure:concept}).
  Additionally, since this deflection varies depending on the weight and length of the grasped tool, we estimate this variation from current sensor information using the mechanism of Parametric Bias (PB) \cite{tani2002parametric} to achieve more accurate control.
  PB is a mechanism that can implicitly embed multiple attractor dynamics within a neural network, and in this study, we utilize it to embed changes in the mutual relationship among sensors.
  Previous studies on learning-based tool-use have explored various methods for tool selection \cite{tee2018tool}, motion planning \cite{xie2019improvisation}, and tool understanding \cite{nabeshima2006tool}, but all of them have been conducted with rigid-bodied robots, and few have considered the center of gravity.
  In the context of imitation learning, methods considering flexible joints exist \cite{wu2018flexible}, but flexible links have not been taken into account.
  Furthermore, since imitation learning requires human demonstrations, its purpose differs from our study, which aims to enable robots to learn the mutual relationship among sensors on their own body.
  Other approaches using reinforcement learning \cite{eppe2019tooluse} or self-supervised learning \cite{fang2020tool} exist, but none of them have considered the flexibility of joints and links, as well as the resulting changes in the body's center of gravity.
  While gradual changes in tool grasping states have been considered in some methods \cite{kawaharazuka2021tooluse}, our study is novel in that it incorporates tactile information and different visual information, considering the weight and length of tools and the resulting body deflection.

  The structure of this study is as follows.
  In \secref{sec:proposed}, we discuss the proposed Whole-body Tool-use Network with Parametric Bias (WTNPB) and the overall system, covering the network structure, network training, online update, and control using this network.
  In \secref{sec:experiments}, we present experimental results using both a simulation and the actual robot, focusing on network training, state estimation based on online update, and tool-tip position control.
  Additionally, we perform an integrated tool-use experiment with the low-rigidity humanoid KXR to open a window.
  In \secref{sec:discussion}, we reflect on the experimental results and outline future prospects.
  In \secref{sec:conclusion}, we provide our concluding remarks.
}%
{%
  そこで本研究では, 低剛性なロボットによる道具利用に向けた, 関節角度や視覚情報, 触覚情報の相互関係を記述するニューラルネットワークを用いた, たわみを考慮した道具先端位置の制御手法を開発する(\figref{figure:concept}).
  また, このたわみは把持した道具の長さや重さによっても変化するため, その変化をParametric Bias \cite{tani2002parametric}の仕組みを応用して現在のセンサ情報から推定し, より正確に制御を行う.
  Parametric Biasは複数のアトラクターダイナミクスをニューラルネットワーク内に暗黙的に埋め込むことが可能な仕組みであり, 本研究ではこれをセンサ間の相互関係変化の埋め込みに利用する.
  学習型での道具利用に関する先行研究は, 道具選択\cite{tee2018tool}や動作計画\cite{xie2019improvisation}, 道具理解\cite{nabeshima2006tool}等について様々な手法が研究されているが, そのどれもが剛な身体での実験であり, また, 重心位置等を考慮したものはほとんどない.
  模倣学習の文脈では柔軟関節を考慮した手法はあるものの\cite{wu2018flexible}, 柔軟リンクは考慮していない.
  また, 模倣学習であるため人間のデモンストレーションが必要な点で, ロボットが自らが身体センサ間の相互関係を学習する本研究の目的とは異なる.
  この他にも強化学習を利用した手法\cite{eppe2019tooluse}や自己教師あり学習を使った手法\cite{fang2020tool}があるものの, 関節やリンクの柔軟性, それに伴う身体重心変化等を考慮した研究はない.
  また, 逐次的な道具把持状態の変化を考慮した手法が開発されている\cite{kawaharazuka2021tooluse}が, 本研究ではこれに接触覚や異なる視覚情報を加え, 物体の重さやそれに伴う身体のたわみを考慮する点で新しい.

  本研究の構成は以下である.
  \secref{sec:proposed}では提案するWhole-body Tool-use Network with Parametric Bias (WTNPB)を含む全体システムについて, ネットワーク構造, ネットワーク訓練, オンライン更新, 本ネットワークを用いた制御について順に述べる.
  \secref{sec:experiments}では, シミュレーションと実機の両者について, ネットワーク訓練, オンライン更新に基づく状態推定, 道具先端位置制御の実験結果を示す.
  また, それらを統合した, 低剛性ロボットKXRによる道具を用いた窓開け実験を行う.
  \secref{sec:discussion}で実験結果と今後の展望について考察し, \secref{sec:conclusion}で結論を述べる.
}%

\section{Whole-body Robotic Tool-use Learning for Low-Rigidity Robots} \label{sec:proposed}
\switchlanguage%
{%
  The overall system, including the Whole-body Tool-use Network with Parametric Bias (WTNPB) proposed in this study, is illustrated in \figref{figure:system-overview}.
}%
{%
  本研究で提案するWhole-body Tool-use Network with Parametric Bias (WTNPB)を含む全体システムを\figref{figure:system-overview}に示す.
}%

\subsection{Network Structure}
\switchlanguage%
{%
  Consider a robot holding a tool in its hand.
  In this scenario, the network structure of WTNPB can be expressed as follows,
  \begin{align}
    \bm{x}^{T} &= \begin{pmatrix}\bm{\theta}^{T}&\bm{x}^{T}_{cog}&\bm{x}^{T}_{tool}&\bm{s}^{T}_{tool}\end{pmatrix}\\
      \bm{z} &= \bm{h}_{enc}(\bm{x}, \bm{m}, \bm{p})\\
      \bm{x} &= \bm{h}_{dec}(\bm{z}) \\
      \bm{h}(\bm{x}, \bm{m}, \bm{p}) &= \bm{h}_{dec}(\bm{h}_{enc}(\bm{x}, \bm{m}, \bm{p})) \label{eq:wtnpb}
  \end{align}
  where $\bm{\theta}$ is the commanded joint angle of the robot, $\bm{x}_{cog}$ is the center of gravity position of the robot, $\bm{x}_{tool}$ is the tool-tip position in 3D space, $\bm{s}_{tool}$ is the tool-tip screen coordinates on the camera image, $\bm{m}$ is the mask variable (will be explained later), $\bm{p}$ is Parametric Bias (PB) \cite{tani2002parametric}, and $\bm{z}$ is the latent variable.
  Furthermore, $\bm{h}_{enc}$ is the encoder network, $\bm{h}_{dec}$ is the decoder network, and $\bm{h}$ is the entire network combining these encoder and decoder networks.

  As detailed in \secref{subsec:exp-setup}, we use a low-rigidity plastic-made humanoid robot, KXR \cite{makabe2021kxr}, for our experiments.
  We define $\bm{\theta}$ as $\begin{pmatrix}\theta_{s-p} & \theta_{s-y} & \theta_{e-p} & \theta_{a-p}\end{pmatrix}^{T}$, representing the commanded joint angle, which include shoulder pitch and yaw angles (s-p, s-y), elbow pitch angle (e-p), and ankle pitch angle (a-p) of KXR.
  we limit the dimensionality of $\bm{\theta}$ to simplify data collection.
  The center of gravity position, $\bm{x}_{cog}=\begin{pmatrix}x_{cog} & y_{cog}\end{pmatrix}^{T}$, is calculated from 1-axis tactile sensors placed at the corners of each foot (assuming that the feet remain aligned during tool-use, and disregarding the $z$-direction).
  $\bm{x}_{tool}$ represents the 3D position of the tool tip recognized using AR markers.
  $\bm{s}_{tool}$ represents the 2D position of the tool tip recognized using AR markers or color extraction.
  The variable $\bm{m}$ is used to capture the relationship among multiple modalities contained in $\bm{x}$.
  Since $\bm{x}$ contains four modalities, $\bm{m}$ is a 4-dimensional vector, where a value of 1 indicates the use of each modality, while 0 indicates its non-use.
  For example, when $\bm{m}^{T} = \begin{pmatrix}1&1&0&0\end{pmatrix}$, the input $\bm{x}^{T}$ consists of $\begin{pmatrix}\bm{\theta}^{T}&\bm{x}^{T}_{cog}&\bm{0}^{T}&\bm{0}^{T}\end{pmatrix}$, meaning that all $\bm{x}$ are reconstructed from a subset of $\bm{x}$ through $\bm{z}$, forming a network structure.
  The choice of $\bm{m}$ is not arbitrary, and it requires defining a feasible set $\mathcal{M}$.
  In this study, as $\bm{\theta}$ serves as both sensor data and commanded values, masks related to $\bm{\theta}$ with a value of $1$ in $\bm{m}$ are all feasible.
  Additionally, we assume that $\bm{\theta}$ can be inferred from all other modalities.
  Therefore, we define $\mathcal{M}$ as $\{(1\;0\;0\;0), (1\;1\;0\;0), (1\;0\;1\;0), (1\;0\;0\;1), (1\;1\;1\;0), (1\;1\;0\;1),\\ (1\;0\;1\;1), (0\;1\;1\;1)\}$.
  This enables the system to adapt to the absence of certain modalities during online update and control.
  Finally, $\bm{p}$ represents PB that captures changes in the mutual relationship among sensors due to changes in the length and weight of the tool.
  In this study, we consider it as a 2D vector.
  Increasing the dimensionality of $\bm{p}$ makes it challenging to structure the space of PB and prone to overfitting, but it allows capturing various changes in the mutual relationship.

  In this study, the overall network structure consists of 7 layers.
  Regarding the number of units, the input is 17-dimensional, composed of 11 dimensions from $\bm{x}$, 4 dimensions from $\bm{m}$, and 2 dimensions from $\bm{p}$.
  The intermediate layers have the following sizes: $\{200, 50, 8, 50, 200\}$, and the output is 11-dimensional corresponding to $\bm{x}$.
  The middle layer with 8 units represents the dimension of the latent variable $\bm{z}$.
  We employ the hyperbolic tangent activation function and the Adam optimizer \cite{kingma2015adam} for update.
  During training, the network input and output are normalized using all collected data.
}%
{%
  あるロボットが手に道具を持った状態を考える.
  このとき, WTNPBのネットワーク構造は以下のように式で表現できる.
  \begin{align}
    \bm{x}^{T} &= \begin{pmatrix}\bm{\theta}^{T}&\bm{x}^{T}_{cog}&\bm{x}^{T}_{tool}&\bm{s}^{T}_{tool}\end{pmatrix}\\
      \bm{z} &= \bm{h}_{enc}(\bm{x}, \bm{m}, \bm{p})\\
      \bm{x} &= \bm{h}_{dec}(\bm{z}) \\
      \bm{h}(\bm{x}, \bm{m}, \bm{p}) &= \bm{h}_{dec}(\bm{h}_{enc}(\bm{x}, \bm{m}, \bm{p})) \label{eq:wtnpb}
  \end{align}
  ここで, $\bm{\theta}$はロボットの指令関節角度, $\bm{x}_{cog}$はロボットの重心位置, $\bm{x}_{tool}$は3次元空間上での道具の先端位置, $\bm{s}_{tool}$はカメラ画像中での道具の先端位置, $\bm{m}$は後に述べるマスク変数, $\bm{p}$はParametric Bias \cite{tani2002parametric}, $\bm{z}$は潜在変数を表す.
  また, $\bm{h}_{enc}$はネットワークのエンコーダ, $\bm{h}_{dec}$はデコーダ, $\bm{h}$はそれらを合わせたネットワーク全体を表す.

  詳細は\secref{subsec:exp-setup}で述べるが, 本研究では低剛性な樹脂製ヒューマノイドKXR \cite{makabe2021kxr}を実験に用いる.
  ここで, $\bm{\theta}=\begin{pmatrix}\theta_{s-p} & \theta_{s-y} & \theta_{e-p} & \theta_{a-p}\end{pmatrix}^{T}$とし, 肩のpitchとyawの角度(s-p, s-y), 肘のpitchの角度(e-p), そして足首のpitchの角度(a-p)の4つを指令関節角度として扱う.
    $\bm{\theta}$の次元が大きくなるほどデータ収集が難しくなるため, 本研究は比較的低次元に制限している.
  重心位置$\bm{x}_{cog}=\begin{pmatrix}x_{cog} & y_{cog}\end{pmatrix}^{T}$はそれぞれの足裏の四隅に配置された1軸の力センサから算出する(道具操作中は足平の位置は変更しないため, 足は揃っているという仮定を置いている, また, $z$方向は無視する).
  $\bm{x}_{tool}$は道具先端位置をARマーカ等により認識した際の, その3次元空間上での位置を表す.
  $\bm{s}_{tool}$は道具先端位置を色認識等によって認識した際の, その画像上の2次元の位置を表す.
  $\bm{m}$は$\bm{x}$に含まれる複数のモダリティの間の相関関係を表現する変数である.
  $\bm{x}$には4つのモダリティが含まれているため, $\bm{m}$は4次元のベクトルであり, 各値について$1$のときはそのモダリティを用い, $0$のときはそのモダリティは用いない.
  例えば$\bm{m}^{T} = \begin{pmatrix}1&1&0&0\end{pmatrix}$のとき, 入力$\bm{x}^{T}$は$\begin{pmatrix}\bm{\theta}^{T}&\bm{x}^{T}_{cog}&\bm{0}^{T}&\bm{0}^{T}\end{pmatrix}$となる.
  つまり, 一部の$\bm{x}$から, $\bm{z}$を介して全ての$\bm{x}$を復元するネットワーク構造となっている.
  $\bm{m}$は何でも良いわけではなく, その集合$\mathcal{M}$を定義する必要がある.
  本研究において$\bm{\theta}$はセンサ値でもあり指令値でもあるため, $\bm{\theta}$の$\bm{m}$が$1$であるマスクは全て実行可能である.
  また, これに追加で, $\bm{\theta}$以外の全てのモーダルから$\bm{\theta}$を推論できると仮定する.
  よって, $\mathcal{M}$は$\{(1\;0\;0\;0), (1\;1\;0\;0), (1\;0\;1\;0), (1\;0\;0\;1),\\(1\;1\;1\;0), (1\;1\;0\;1), (1\;0\;1\;1), (0\;1\;1\;1)\}$とした.
  これにより, オンライン学習や制御の際に, 一部のモダリティの欠損に対応できるようになる.
  $\bm{p}$は道具の長さや重さの違いによるダイナミクスの変化を表現するParametric Biasであり, 本研究では2次元としている.
  $\bm{p}$の次元は大きくするほど, $\bm{p}$の空間を構造化することが難しく過学習しやすくなる一方で, 多様な相関関係の変化を捉えられるようになる.

  本研究では, 全体のネットワーク構造は7層とし, ユニット数については, 入力は$\bm{x}$の11次元, $\bm{m}$の4次元, $\bm{p}$の2次元を足し合わせた17次元, 中間層は$\{200, 50, 8, 50, 200\}$, 出力は$\bm{x}$の11次元とした.
  この中間層のユニット数8の部分が潜在変数$\bm{z}$の次元である.
  活性化関数はtanh, 更新則はAdam \cite{kingma2015adam}とする.
  ネットワークの入力と出力は訓練時に得られたデータを使い正規化されている.
}%

\subsection{Training of WTNPB}
\switchlanguage%
{%
  First, we collect data for various tool states (weight and length), denoted as $k$ ($1 \leq k \leq K$, where $K$ is the total number of tool states used for training), in different poses, represented as $D_{k}=\{\bm{x}_{1}, \cdots, \bm{x}_{N_{k}}\}$, where $N_{k}$ is the number of data points for tool state $k$.
  Additionally, we prepare Parametric Bias $\bm{p}_k$ as a learnable input variable for each tool state $k$ (all $\bm{p}_{k}$ are initialized to 0).
  It is important to note that there is no need to explicitly know the specific weight or length of the tools, as the space of PB self-organizes implicitly.
  We generate the dataset $D_{train} = \{(D_{1}, \bm{p}_{1}), \cdots, (D_{N_{k}}, \bm{p}_{N_k})\}$ and train $\bm{h}$ using this dataset.
  During training, $\bm{p}_{k}$ is common for data $D_{k}$ but varies across different tool states.
  The training process involves updating the network weight $W$ and $\bm{p}_k$ ($1 \leq k \leq K$) simultaneously using backpropagation.
  Furthermore, we randomly select $\bm{m}$ from $\mathcal{M}$ as a network input.
  This mechanism allows information about tool state $k$ to be embedded in each $\bm{p}_{k}$, while the space of $\bm{p}$ self-organizes based on the tool state.

  The training process consists of two stages.
  First, we vary the tool state in a simulation to collect data and calculate $W$ and $\bm{p}_{k}$.
  In this stage, we generate data by randomly moving $\bm{\theta}$ within a specified joint angle range.
  $\bm{s}_{tool}$ and $\bm{x}_{cog}$ are computed using camera models and kinematic models.
  For low-rigidity robots, we make a simple model for joint error based on joint torque to obtain data, which assumes an angular deflection of 3.0$\tau$ ($\tau$ is joint torque [Nm]).
  Subsequently, we retain only the computed $W$ from the simulation and initialize $\bm{p}_{k}$ to $\bm{0}$.
  Second, we collect data using the actual robot and perform fine-tuning.
  To facilitate data collection, AR markers are attached to the tool tip, allowing us to collect all values of $\bm{x}$, including $\bm{x}_{tool}$ and $\bm{s}_{tool}$.
  We employ two methods for data collection.
  One involves direct human teaching of $\bm{\theta}$ via a GUI, with data collected for each specified pose.
  The other method randomly decides $\bm{\theta}$ in a simulation, while ensuring that the center of gravity remains within the support area and the tool-tip position is visible from the camera.
  By imposing strong constraints on the center of gravity position, we can collect data safely within the range of stability, even if there are differences between a simulation and the actual robot.
  Since the actual robot data is limited, fine-tuning is performed using data from a simulation to adapt to real-world conditions.
}%
{%
  まず, 道具の長さや重さが異なるような, 様々な道具状態$k$ ($1 \leq k \leq K$, $K$は訓練に使う全道具状態数)について, 様々な姿勢におけるデータ$D_{k}=\{\bm{x}_{1}, \cdots, \bm{x}_{N_{k}}\}$を収集する($N_{k}$は道具状態$k$に関するデータ数).
  また, それぞれの道具状態$k$について学習可能な入力変数であるParametric Bias $\bm{p}_k$を用意する(全ての$\bm{p}_k$は0に初期化されている).
  このとき, 具体的な道具の長さや重さを知る必要はなく, PBは暗黙的に自己組織化する.
  データ$D_{train} = \{(D_{1}, \bm{p}_{1}), \cdots, (D_{N_{k}}, \bm{p}_{N_k})\}$が収集され, この$D_{train}$を用いて$\bm{h}$を学習させる.
  このとき, $\bm{p}_{k}$はデータ$D_{k}$については共通であり, 異なる道具状態については異なる変数である.
  学習の際はネットワークの重み$W$と$\bm{p}_k$ ($1 \leq k \leq K$)を同時に誤差逆伝播法によって更新する.
  また, $\bm{m}$は$\mathcal{M}$からランダムに選び入力とする.
  これにより, 各$\bm{p}_{k}$に道具状態$k$に関する情報が埋め込まれるとともに, $\bm{p}$の空間が道具状態によって自己組織化される.

  本研究では学習は2段階で行う.
  まず, シミュレーション上で道具状態を変化させてデータを収集し, $W$と$\bm{p}_{k}$を計算する.
  ここでは, $\bm{\theta}$を設定した関節角度範囲内でランダムに動かすことでデータを収集する.
  $\bm{s}_{tool}$や$\bm{x}_{cog}$については, カメラやリンク重量等のモデルから計算する.
  関節トルクに応じた関節誤差を簡易的にモデル化し, $3.0\tau$ ($\tau$の単位は[Nm])だけ関節がたわむ(関節角度が変化する)と仮定してデータを取得している.
  その後, シミュレーションで計算された$W$のみ残し, $\bm{p}_{k}$を0に初期化する.
  最後に, 実機においてデータを収集し, 学習を行う.
  なお, 道具先端にARマーカをつけることで, $\bm{x}_{tool}$と$\bm{s}_{tool}$を含めた$\bm{x}$の全ての値をデータとして収集する.
  ここでは, ニ種類の方法でデータを収集する.
  一つはGUIから人間が$\bm{\theta}$を直接指定し, そのときのデータを収集していく方法である.
  もう一つはシミュレーションにおいて$\bm{\theta}$をランダムに指定するが, このとき重心が支持領域内に含まれ, かつ道具先端位置がカメラから見える場合の$\bm{\theta}$を実機に送って動かす.
  重心位置に関する制約を強くすることで, たわみによりシミュレーションと実機が異なっても, 倒れない範囲でデータを収集することが可能である.
  実機で得られるデータは少数なため, シミュレーションからのFine Tuningにより対応している.
}%

\subsection{Online Update of Parametric Bias}
\switchlanguage%
{%
  When grasping a new tool, it is necessary to update information about its length and weight.
  Therefore, data is obtained by moving the robot body, and online update of Parametric Bias $\bm{p}$ is performed.
  For any value $\bm{x}_{i}$ in $\bm{x}$, if it can be obtained and is sufficiently different from the previously collected value ($||\bm{x}_{i}-\bm{x}^{prev}_{i}||_{2}>C^{online}_{i}$, where $\bm{x}^{prev}_{i}$ is the previously collected $\bm{x}_{i}$ and $C^{online}_{i}$ is a threshold), all values of $\bm{x}$ that can be obtained at that moment are collected as data.
  The online update begins when the obtained data count $N^{online}$ exceeds a threshold $N^{online}_{thre}$ and continues with each subsequent data collection.
  The weight $W$ is kept fixed, and only $\bm{p}$ is updated with a batch size of $N^{online}_{batch}$ and a number of epochs of $N^{online}_{epoch}$.
  The update rule uses Momentum SGD with a learning rate of 0.01.
  Data is limited to a maximum of $N^{online}_{max}$ ($N^{online}_{thre} \leq N^{online}_{max}$), and data exceeding this limit is removed in a first-in-first-out manner ($N^{online} \leq N^{online}_{max}$).
  By keeping the network weight $W$ fixed and updating only the low-dimensional PB, overfitting is prevented while updating only the tool state.
  It should be noted that a mask $\bm{m}$, which indicates $1$ for all the obtained data in $\bm{x}$, must be included in the set of feasible masks $\mathcal{M}$.
  Using this $\bm{m}$, $\bm{z}$ is calculated, and losses are computed only for the obtained data in the output $\bm{x}$.
  The more data we obtain for $\bm{x}$, the better the online update should work.
  In this study, the tool-tip position $\bm{x}_{tool}$ is obtained using AR markers only during the training data collection.
  During online update, the AR marker obstructs tool manipulation, so the tool tip is painted red for color recognition and acquisition of $\bm{s}_{tool}$ , but $\bm{x}_{tool}$ cannot be obtained.
  Therefore, $\bm{p}$ is updated from $\{\bm{\theta}, \bm{x}_{cog}, \bm{s}_{tool}\}$.

  In this study, we set $C_{collect}=\{10$ [deg], $3$ [mm], $20$ [mm], $100$ [px]$\}$, $N^{online}_{thre}=5$, $N^{online}_{batch}=N^{online}$, $N^{online}_{epoch}=5$, and $N^{online}_{max}=100$.
  Additionally, the data collection is performed at 5 Hz.
}%
{%
  新しく道具を持った際に, この道具の長さや重さに関する情報を更新する必要がある.
  そのため, 自身の身体を少し動かすことでデータを取得し, Parametric Bias $\bm{p}$のオンライン更新を行う.
  $\bm{x}$のいずれかの値について, その値が取得できるかつ, 直前に収集された値よりもある程度現在の値が離れている場合($||\bm{x}_{i}-\bm{x}^{prev}_{i}||_{2}>C^{online}_{i}$, $\bm{x}^{prev}_{i}$は直前に収集された$\bm{x}_{i}$, $C^{online}_{i}$は閾値)に, そのタイミングで取得できている$\bm{x}$の全ての値をデータとして収集する.
  得られたデータ数$N^{online}$が, $N^{online}_{thre}$を超えてから更新を開始し, その後新しいデータ収集される度に更新を行う.
  重み$W$を固定し, $\bm{p}$のみをバッチ数を$N^{online}_{batch}$, エポック数を$N^{online}_{epoch}$として更新する.
  この際の更新則は, 学習率を0.01としたMomentum SGDとする.
  データは$N^{online}_{max}$ ($N^{online}_{thre} \leq N^{online}_{max}$)を最大値とし, それを超えたデータは古いものから順に削除していく($N^{online} \leq N^{online}_{max}$).
  ネットワークの重み$W$を固定し, 小さな次元であるParametric Biasのみ更新することで, 過学習を防ぎつつ道具状態のみを更新することができる.
  なお, このとき$\bm{x}$のうち得られたデータのみを使うマスク$\bm{m}$が, 実行可能なマスク集合$\mathcal{M}$に含まれている必要がある.
  この$\bm{m}$を使って$\bm{z}$を計算し, 出力した$\bm{x}$に対して, 得られたデータについてのみ損失を計算する.
  $\bm{x}$のうち得られるデータが多ければ多いほどオンライン更新はうまく動作するはずである.
  本研究では訓練データ収集時のみ道具先端位置$\bm{x}_{tool}$をARマーカを使って取得する.
  このオンライン更新時にはARマーカが道具操作の邪魔となるため, 色認識に向けて道具の先端を赤く塗るのみとしており, $\bm{x}_{tool}$は得られない.
  よって, $\{\bm{\theta}, \bm{x}_{cog}, \bm{s}_{tool}\}$から$\bm{p}$を更新する.

  本研究では, $C_{collect}=\{10$ [deg], $3$ [mm], $20$ [mm], $100$ [px]$\}$, $N^{online}_{thre}=5$, $N^{online}_{batch}=N^{online}$, $N^{online}_{epoch}=5$, $N^{online}_{max}=100$とした.
  また, データ収集の実行周期は5 Hzである.
}%

\subsection{Control using WTNPB}
\switchlanguage%
{%
  The control of the tool-tip position is achieved through optimization using backpropagation and gradient descent.
  The optimization process is as follows,
  \begin{align}
    L &= ||\bm{x}^{pred}_{tool}-\bm{x}^{ref}_{tool}||_{2} +  \alpha||\bm{x}^{pred}_{cog}-\bm{x}^{ref}_{cog}||_{2} \label{eq:control-loss}\\
    \bm{z}^{opt} &\gets \bm{z}^{opt} + \gamma\partial{L}/\partial{\bm{z}^{opt}} \label{eq:control-opt}
  \end{align}
  where $\bm{x}^{ref}_{tool}$ is the commanded tool-tip position, $\bm{x}^{ref}_{cog}$ is the commanded center of gravity position (in this study, $\bm{x}^{ref}_{cog}=\begin{pmatrix}0&0\end{pmatrix}^{T}$), $\bm{z}^{opt}$ is the value of $\bm{z}$ being optimized, $\bm{x}^{pred}_{\{tool, cog\}}$ is the predicted $\bm{x}_{\{tool, cog\}}$ derived from $\bm{z}^{opt}$, $\alpha$ is the weight of the loss function, and $\gamma$ is the learning rate.
  In essence, this control aims to bring the tool-tip position closer to the commanded value while maintaining the center of gravity position at the center of the foot.
  The initial value of $\bm{z}^{opt}$ is obtained using predictions from $\bm{x}^{ref}_{\{tool, cog\}}$ and the corresponding $\bm{m}$ through $\bm{h}_{enc}$.
  We update $\bm{z}^{opt}$ from the loss $L$ and send $\bm{\theta}$ computed from the final $\bm{z}^{opt}$ to the robot.
  Regarding $\gamma$, a set of $N^{control}_{batch}$ $\gamma$ values, exponentially divided within the range $[0, \gamma_{max}]$, is prepared.
  For each $\gamma$, \equref{eq:control-opt} is executed, followed by loss calculation using \equref{eq:control-loss}.
  This process is repeated for $N^{control}_{epoch}$ iterations, and $\bm{z}^{opt}$ with the smallest loss is selected.
  By experimenting with various $\gamma$ values and selecting the best learning rate, faster convergence can be achieved.
  Additionally, we can incorporate a term like $||\bm{\theta}^{pred}-\bm{\theta}^{cur}||_{2}$ (where $\bm{\theta}^{\{pred, cur\}}$ is the predicted or current value of $\bm{\theta}$) into the cost function $L$ if we wish to minimize the movement of joint angle from the current value, or $||\bm{s}^{pred}_{tool}-\bm{s}^{ref}_{tool}||_{2}$ (where $\bm{s}^{\{pred, ref\}}_{tool}$ is the predicted or commanded value of $\bm{s}_{tool}$) if the desired tool-tip screen coordinates on the camera image are obtained.

  In this study, we set $\alpha=0.01$, $\gamma_{max}=0.1$, $N^{control}_{batch}=30$, and $N^{control}_{epoch}=30$.
}%
{%
  道具先端位置の制御は誤差逆伝播と勾配降下法を使った最適化により行う.
  以下のように最適化を行う.
  \begin{align}
    L &= ||\bm{x}^{pred}_{tool}-\bm{x}^{ref}_{tool}||_{2} +  \alpha||\bm{x}^{pred}_{cog}-\bm{x}^{ref}_{cog}||_{2} \label{eq:control-loss}\\
    \bm{z}^{opt} &\gets \bm{z}^{opt} + \gamma\partial{L}/\partial{\bm{z}^{opt}} \label{eq:control-opt}
  \end{align}
  ここで, $\bm{x}^{ref}_{tool}$は道具先端位置の指令値, $\bm{x}^{ref}_{cog}$は指令重心位置(本研究では$\begin{pmatrix}0&0\end{pmatrix}^{T}$), $\bm{z}^{opt}$は最適化する$\bm{z}$の値, $\bm{x}^{pred}_{\{tool, cog\}}$は$\bm{z}^{opt}$から予測された$\bm{x}_{\{tool, cog\}}$, $\alpha$は損失関数の重み, $\gamma$は学習率を表す.
  つまり, 道具先端位置を指令値に近づけつつ, 重心位置を足裏中心に保つように制御する.
  $\bm{z}^{opt}$の初期値は$\bm{m}=\begin{pmatrix}0&1&0&1\end{pmatrix}^{T}$として$\bm{x}^{ref}_{\{tool, cog\}}$と$\bm{h}_{enc}$から予測された値を用いる.
  $\bm{z}^{opt}$を損失$L$から更新していき, 最終的に計算された$\bm{z}^{opt}$から得られた$\bm{\theta}$を実機に送る.
  なお, $\gamma$については$[0, \gamma_{max}]$を指数関数的に分割した$N^{control}_{batch}$個の$\gamma$を用意し, それぞれの$\gamma$で\equref{eq:control-opt}を実行した後, \equref{eq:control-loss}で損失を計算し最も損失の小さい$\bm{z}^{opt}$を選択することを$N^{control}_{epoch}$回繰り返す.
  様々な$\gamma$を試して最良の学習率を常に選ぶことで, より速い収束が得られる.
  また, もし指令関節角度を現在の値からなるべく動かしたくなければ$||\bm{\theta}^{pred}-\bm{\theta}^{cur}||_{2}$ ($\bm{\theta}^{\{pred, cur\}}$は$\bm{\theta}$の予測値または現在値), カメラ画像中での道具の指令先端位置を得られる場合は$||\bm{s}^{pred}_{tool}-\bm{s}^{ref}_{tool}||_{2}$ ($\bm{s}^{\{pred, ref\}_{tool}}$は$\bm{s}_{tool}$の予測値または指令値)などを$L$に加えても良い.

  本研究では, $\alpha=0.01$, $\gamma_{max}=0.1$, $N^{control}_{batch}=30$, $N^{control}_{epoch}=30$とした.
}%

\section{Experiments} \label{sec:experiments}

\begin{figure}[t]
  \centering
  \includegraphics[width=0.8\columnwidth]{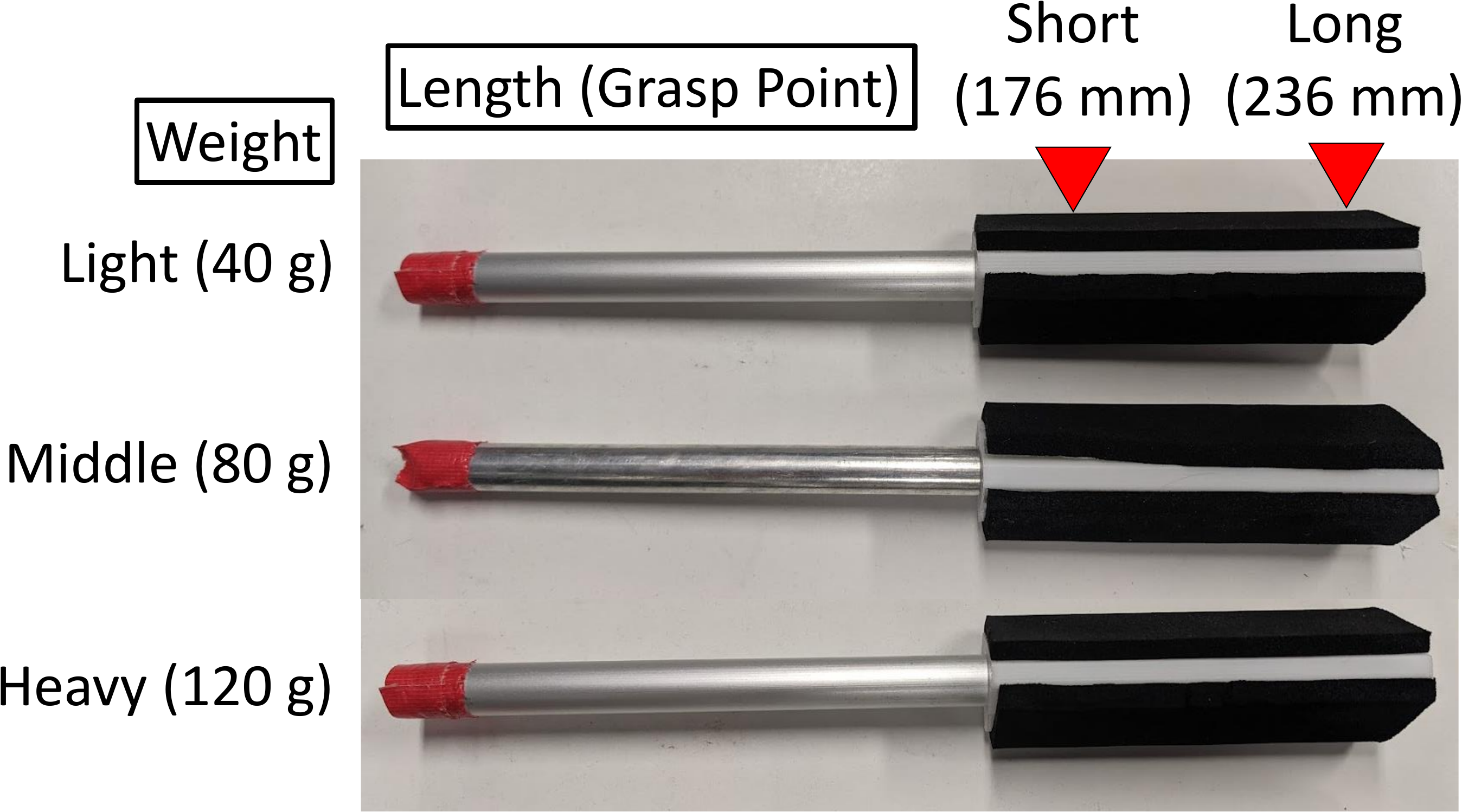}
  \vspace{-1.0ex}
  \caption{Six types of tool states with various weights and lengths used in this study.}
  \label{figure:experimental-setup}
  \vspace{-1.0ex}
\end{figure}

\begin{figure}[t]
  \centering
  \includegraphics[width=0.8\columnwidth]{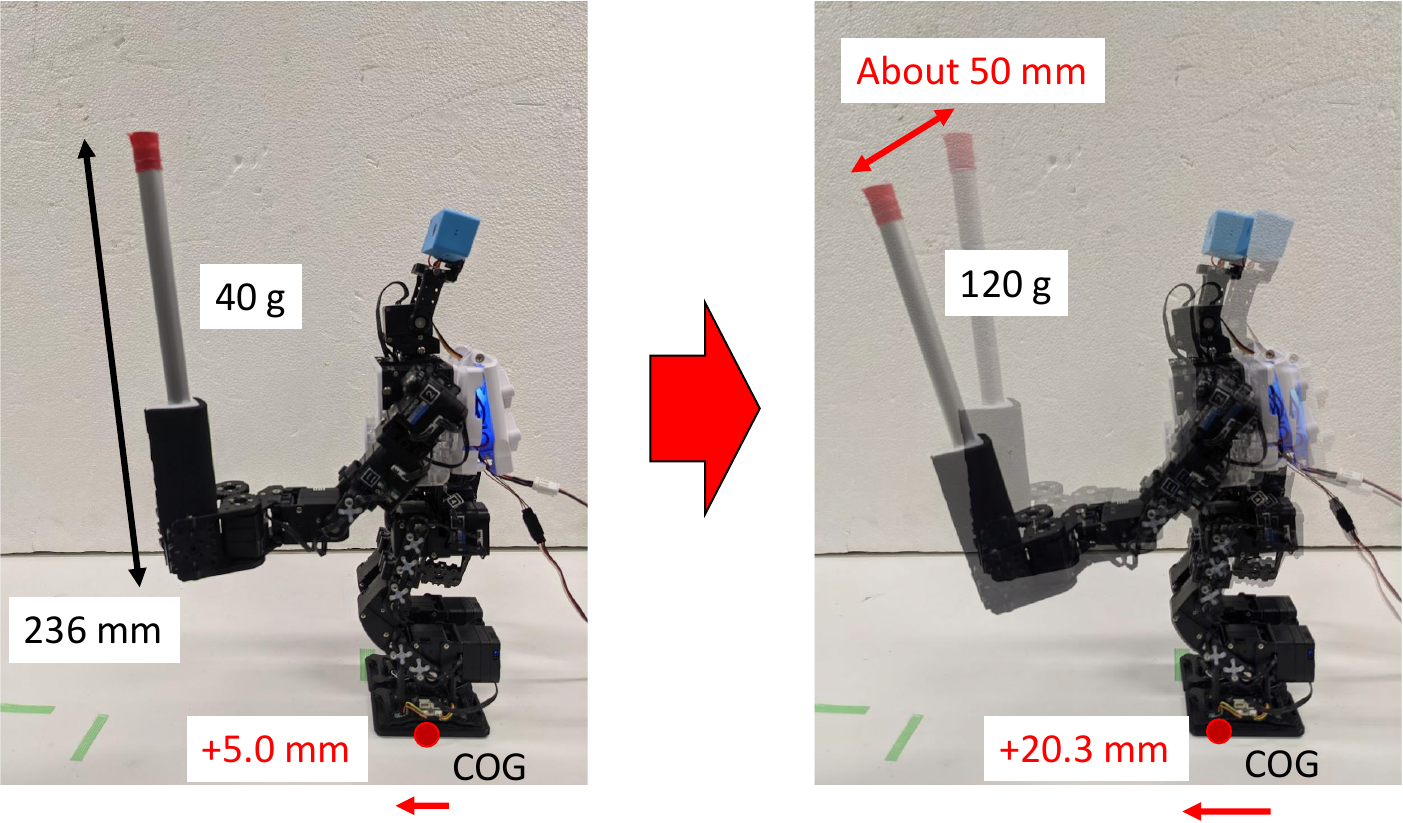}
  \vspace{-1.0ex}
  \caption{The change in tool-tip position and center of gravity when handling tools with different weights.}
  \label{figure:flex-feature}
  \vspace{-1.0ex}
\end{figure}

\subsection{Experimental Setup} \label{subsec:exp-setup}
\switchlanguage%
{%
  In this study, we utilized a low-rigidity humanoid robot, KXR \cite{makabe2021kxr}, made of plastic, as shown in \figref{figure:concept}.
  Three different tool weights (Light: 40 g, Middle: 80 g, Heavy: 120 g) were prepared, as depicted in \figref{figure:experimental-setup}.
  Additionally, based on the grasped positions of these tools, we defined two lengths (Short: 176 mm, Long: 236 mm), resulting in a total of six tool states.
  In \figref{figure:flex-feature}, we illustrate the variations in the tool-tip position and center of gravity (COG) when holding tools of different weights.
  Even when using tools of the same length, due to the low rigidity of the hardware, there is a significant difference in the tool-tip position (about 50 mm) and the center of gravity position (about 15 mm) when holding a 40 g tool compared to a 120 g tool.
  Similarly, when the tool length changes, both the tool-tip position $\bm{x}_{tool}$ and screen coordinates $\bm{s}_{tool}$, as well as the center of gravity position $\bm{x}_{cog}$, change accordingly.
}%
{%
  本研究では\figref{figure:concept}に示す低剛性な樹脂製ヒューマノイドKXR \cite{makabe2021kxr}を実験に用いる.
  \figref{figure:experimental-setup}に示すように, 3種類の重さ(Light: 40 g, Middle: 80 g, Heavy: 120 g)の道具を用意する.
  また, それらの道具の持つ位置に応じて, 2種類の長さ(Short: 全長176 mm, Long: 全長236mm)を定義し, これらの組み合わせである計6種類の道具状態を扱う.
  \figref{figure:flex-feature}に, 異なる重さの道具を持った際の道具先端位置と重心位置の変化を示す.
  同じ長さの道具でも, ハードウェアが低剛性であるがゆえに, 40 gの道具と120 gの道具を持った時では道具先端位置は約50 mm, 重心位置は約15mmも異なる.
  同様に, 道具の長さが変わればその道具先端位置$\bm{x}_{tool}$や$\bm{s}_{tool}$が変化すると同時に, 重心位置$\bm{x}_{cog}$も変化する.
}%

\begin{figure}[t]
  \centering
  \includegraphics[width=0.8\columnwidth]{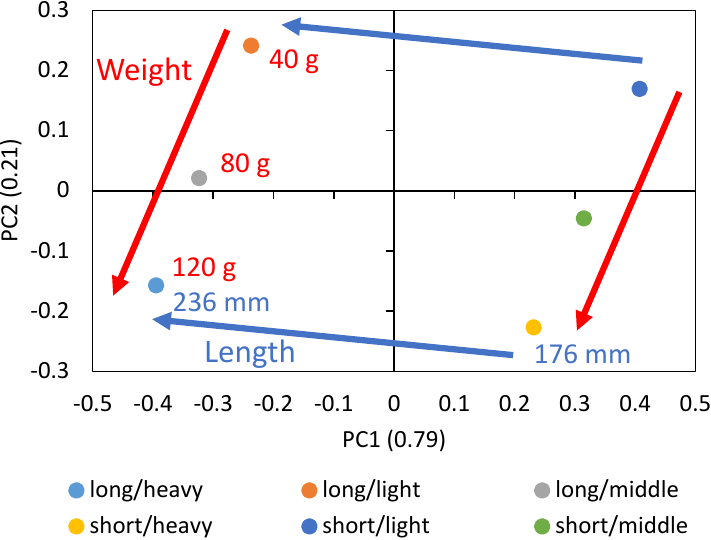}
  \vspace{-1.0ex}
  \caption{The trained parametric bias in the simulation experiment.}
  \label{figure:sim-pb}
  \vspace{-1.0ex}
\end{figure}

\begin{figure}[t]
  \centering
  \includegraphics[width=0.8\columnwidth]{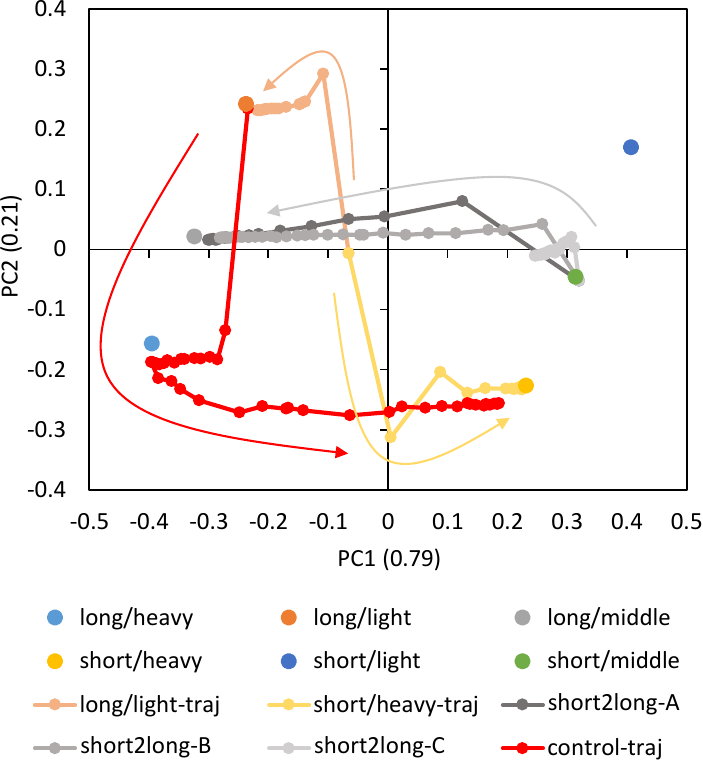}
  \vspace{-1.0ex}
  \caption{The trajectory of parametric bias during online update in the simulation experiment.}
  \label{figure:sim-online}
  \vspace{-1.0ex}
\end{figure}

\begin{figure}[t]
  \centering
  \includegraphics[width=0.8\columnwidth]{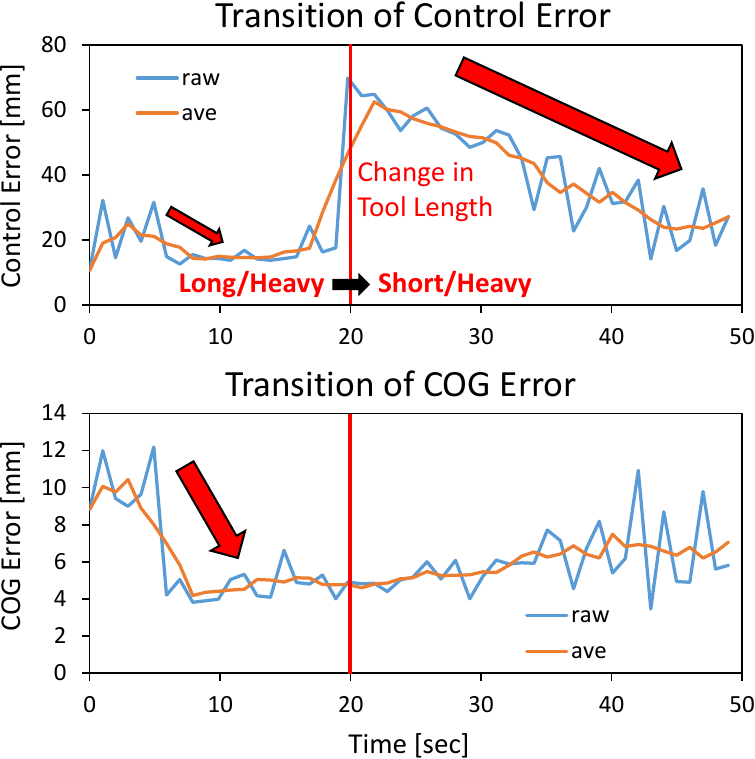}
  \vspace{-1.0ex}
  \caption{The transition of control and center of gravity errors in the simulation experiment.}
  \label{figure:sim-control}
  \vspace{-1.0ex}
\end{figure}

\subsection{Simulation Experiment}
\switchlanguage%
{%
  The joint angle was randomly varied in a simulation, resulting in 500 data points for each tool state and a total of 3000 data points.
  The results of applying Principal Component Analysis (PCA) to PB obtained when training WTNPB using this data are shown in \figref{figure:sim-pb}.
  It can be observed that each PB aligns neatly along the axes of tool weight and length.

  Next, we demonstrate the accurate recognition of the current tool state through online update of PB.
  The tool was set to Long/Light or Short/Heavy states, and PB was updated online while sending random joint angles to the robot.
  The transitions of PB during this process are illustrated in \figref{figure:sim-online} (long/light-traj and short/heavy-traj).
  It can be observed that the current PB values gradually approach the trained PB values for Long/Light or Short/Heavy.
  Additionally, we consider how online update is affected by data from different types of sensors obtained during operation.
  Regarding visual information, when an AR marker is attached to the tool tip, $\{\bm{x}_{tool}, \bm{s}_{tool}\}$ can be obtained, but without an AR marker, only $\bm{s}_{tool}$ can be obtained based on color recognition.
  Furthermore, if the tool tip is not within the visual field, only the values $\{\bm{\theta}, \bm{x}_{cog}\}$ can be obtained.
  Therefore, in this study, we examine how PB values transition in three scenarios of data availability, A: $\{\bm{\theta}, \bm{x}_{cog}, \bm{x}_{tool}, \bm{s}_{tool}\}$, B: $\{\bm{\theta}, \bm{x}_{cog}, \bm{s}_{tool}\}$, and C: $\{\bm{\theta}, \bm{x}_{cog}\}$.
  Here, we start from the trained PB for Short/Middle and investigate the case when handling the Long/Middle tool.
  The results are presented in \figref{figure:sim-online}.
  A (short2long-A) uses the same sensor types as the earlier long/light-traj and short/heavy-traj scenarios, so the PB quickly transitions and the tool state is accurately recognized.
  B (short2long-B) exhibits a slower transition compared to A, requiring more time to accurately perceive the tool state.
  C (short2long-C) shows that while the PB value is progressing in the correct direction, it has not reached the point of accurate recognition.

  Finally, we conduct control experiments incorporating online update of PB.
  We initiate the current PB as Long/Light, and sequentially transition the actual tool states to Long/Heavy and Short/Heavy.
  It should be noted that in order to promptly adapt to tool changes, we set $N^{online}_{max}=5$.
  Random $\bm{x}^{ref}_{tool}$ values within the specified range are provided, and tool-tip control using WTNPB is executed to track these references.
  The transition of PB during this process is shown in \figref{figure:sim-online} (control-traj).
  It can be observed that PB transitions from Long/Light to Long/Heavy and then to Short/Heavy as intended.
  The control error $||\bm{x}^{ref}_{tool}-\bm{x}_{tool}||_{2}$, center of gravity (COG) error $||\bm{x}^{ref}_{cog}-\bm{x}_{cog}||_{2}$, and their moving averages over five steps are shown in \figref{figure:sim-control}.
  Initially in the Long/Heavy state, we can see a slight reduction in control error and a significant decrease in COG error by the update of PB.
  In the subsequent transition to the Short/Heavy state, significant changes in control error are observed, while there is no significant variation in COG error.
}%
{%
  シミュレーションにおいて関節角度をランダムに動かし, 一つの道具状態に対して500, 全体で3000のデータを取得した.
  このデータを用いてWTNPBを学習した際に得られたParametric Biasに対してPrinciple Component Analysis (PCA)をかけ, 2次元平面上にプロットしたものを\figref{figure:sim-pb}に示す.
  それぞれのParametric Biasが, 道具の重さと長さの軸に沿って綺麗に自己組織化していることがわかる.

  次に, このParametric Biasをオンライン更新することで, 現在の道具状態を正確に認識できることを示す.
  道具をLong/Lightの状態, Short/Heavyの状態に設定し, ランダムな関節角度を送りながらParametric Biasのオンライン更新を実行した.
  その際のParametric Biasの遷移を\figref{figure:sim-online}に示す(long/light-trajとshort/heavy-traj).
  現在のPB値が, 訓練時のLong/Light, またはShort/HeavyのPB値へと徐々に近づいていることがわかる.
  また, 動作時に得られるセンサの種類の違いにより, どうオンライン更新が変化するかを考える.
  道具先端にARマーカがついていれば$\{\bm{x}_{tool}, \bm{s}_{tool}\}$が得られるが, ARマーカなしで色認識のみの場合は$\bm{s}_{tool}$しか得られない.
  また, 視覚に道具先端が入らない場合は, $\{\bm{\theta}, \bm{x}_{cog}\}$の値しか得られない.
  そこで, 本研究では得られるデータがA: $\{\bm{\theta}, \bm{x}_{cog}\}$, B: $\{\bm{\theta}, \bm{x}_{cog}, \bm{s}_{tool}\}$, C: $\{\bm{\theta}, \bm{x}_{cog}, \bm{x}_{tool}, \bm{s}_{tool}\}$の3種類の場合で, どのようにPB値が遷移するかを検証する.
  ここでは, Short/MiddleにおけるPBからスタートし, Long/Middleの道具を扱った場合について検証する.
  その結果を同様に\figref{figure:sim-online}に示す.
  C (short2long-C)は前述のlong/light-trajやshort/heavy-trajとセンサの種類が同じであり, 素早くPBが遷移し, 正確に道具状態を把握できている.
  B (short2long-B)はCよりも遷移のスピードが落ち, 正確に道具状態を把握するのに時間を要している.
  A (short2long-A)は正しい方向にPB値自体は進んでいるものの, 正しくPBを認識するところまでには至っていない.

  最後に, Parametric Biasのオンライン更新を含んだ制御実験を行う.
  道具状態をLong/Lightとして認識している状態から始め, 実際の道具状態をLong/Heavy, Short/Heavyの順で変更する.
  なお, ここでは道具変化にすぐ対応できるように, $N^{online}_{max}=5$とした.
  指定した範囲内のランダムな$\bm{x}^{ref}_{tool}$を与え, これに追従するよう, 前述のWTNPBを使った動作制御を実行する.
  このときのParametric Biasの遷移を\figref{figure:sim-online}のcontrol-trajに示す.
  Long/LightからLong/Heavy, Short/Heavyの順にParametric Biasが遷移していることがわかる.
  また, このときの制御誤差$||\bm{x}^{ref}_{tool}-\bm{x}_{tool}||_{2}$と重心位置誤差$||\bm{x}^{ref}_{cog}-\bm{x}_{cog}||_{2}$, それぞれの5ステップ分の平均を\figref{figure:sim-control}に示す.
  まず, Long/Heavyの状態では, PB値が更新されることで, 制御誤差が少し, また, 重心位置誤差が大きく減少していることがわかる.
  PB値がLong/LightからLong/Heavyに変化し, 道具の長さは変わらないが重さが大きく変化したため, 制御誤差に比べて重心位置誤差が大きく減少したのだと考えられる.
  その後, Short/Heavyの状態に移行した際には, 大きく制御誤差が変化するが, 重心位置誤差に大きな変化はない.
}%

\begin{figure}[t]
  \centering
  \includegraphics[width=0.8\columnwidth]{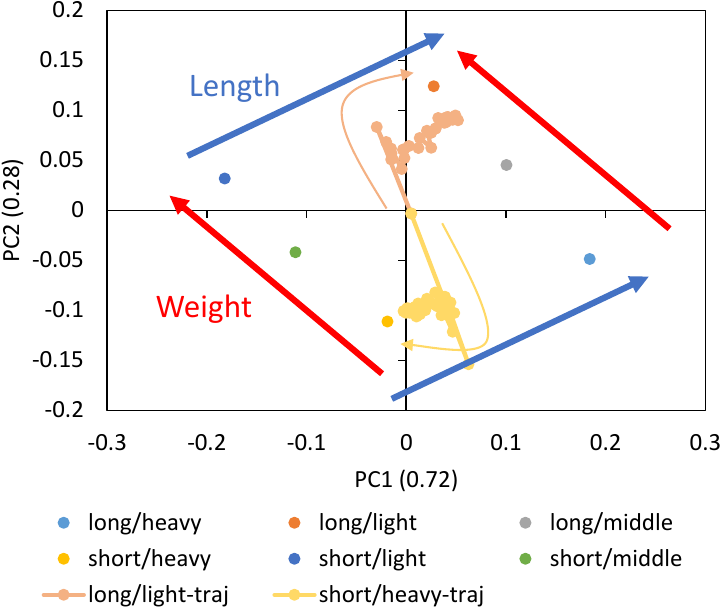}
  \vspace{-1.0ex}
  \caption{The trained parametric bias and the trajectory of the parametric bias during online update in the actual robot experiment.}
  \label{figure:act-pb}
  \vspace{-1.0ex}
\end{figure}

\begin{figure}[t]
  \centering
  \includegraphics[width=0.8\columnwidth]{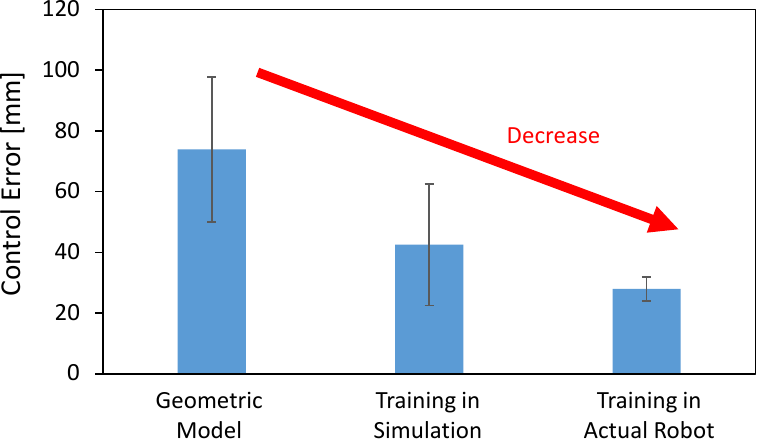}
  \vspace{-1.0ex}
  \caption{Comparison of control errors among the cases using a geometric model, using WTNPB after training with the simulation data, and using WTNPB after fine-tuning with the actual robot data, in the actual robot experiment.}
  \label{figure:act-control}
  \vspace{-1.0ex}
\end{figure}

\begin{figure*}[t]
  \centering
  \includegraphics[width=1.9\columnwidth]{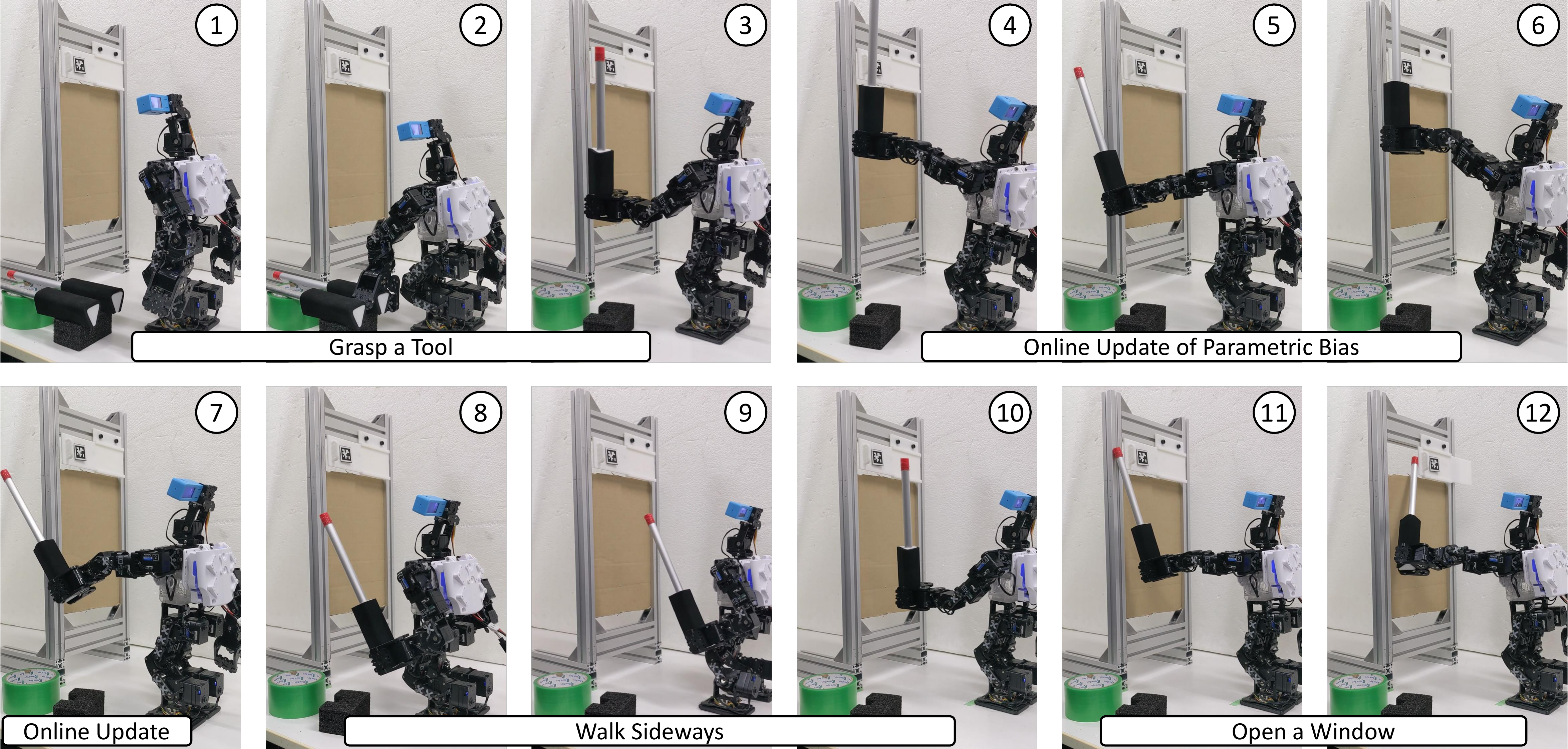}
  \vspace{-1.0ex}
  \caption{The integrated experiment of opening a window.}
  \label{figure:integrated-exp}
  \vspace{-1.0ex}
\end{figure*}

\subsection{Actual Robot Experiment}
\switchlanguage%
{%
  In this study, approximately 480 data points were obtained by collecting data with a GUI interface (60 data points) and with randomly specified joint angles (20 data points) while changing the tool states.
  The results of applying PCA to PB obtained during fine-tuning using this data are shown in \figref{figure:act-pb}.
  It is evident that each PB is self-organized neatly along the axes of tool weight and length.

  Next, we demonstrate the ability to accurately recognize the current tool state through online update of PB.
  PB was updated online while setting the tool state to Long/Light or Short/Heavy, and sending random joint angles to the robot.
  The transitions of PB during this process are also shown in \figref{figure:act-pb} (long/light-traj and short/heavy-traj).
  It can be observed that the current PB gradually approaches the trained PB of Long/Light or Short/Heavy.

  Finally, an evaluation of control errors was conducted.
  We compared control errors for the Long/Middle tool state when solving whole-body inverse kinematics using a geometric model, when using WTNPB trained with the simulation data including joint deflection, and when using WTNPB fine-tuned with the actual robot data.
  The results comparing the average and variance of control errors for five $\bm{x}^{ref}_{tool}$ are shown in \figref{figure:act-control}.
  It should be noted that PB used with WTNPB is the value obtained during training for Long/Middle.
  The geometric model exhibits the largest error, followed by training from the simulation data with joint deflection, and finally, the smallest control error is achieved after fine-tuning in the actual robot.
}%
{%
  実機においてGUIを用いたデータ取得(60データ)と関節角度のランダム指定を用いたデータ取得(20データ)により, 道具を変えながら全部で約480のデータを取得した.
  このデータを用いたfine tuningの際に得られたParametric Biasに対して, PCAをかけた結果を\figref{figure:act-pb}に示す.
  それぞれのParametric Biasが, 道具の重さと長さの軸に沿って綺麗に自己組織化していることがわかる.

  次に, このParametric Biasをオンライン更新することで, 現在の道具状態を正確に認識できることを示す.
  道具をLong/Lightの状態, Short/Heavyの状態に設定し, ランダムな関節角度を送りながらParametric Biasのオンライン更新を実行した.
  その際のParametric Biasの遷移を同様に\figref{figure:act-pb}に示す(long/light-trajとshort/heavy-traj).
  現在のPB値が, 訓練時のLong/Light, またはShort/HeavyのPB値へと徐々に近づいていることがわかる.

  最後に, 制御誤差に関する評価を行う.
  Long/Middleの道具状態において, 幾何モデルを用いて全身逆運動学を解いた場合, 関節のたわみを含むシミュレーションデータからWTNPBを学習させた場合, これを実機においてFine Tuningした場合について比較実験を行う.
  5つの$\bm{x}^{ref}_{tool}$に関する制御誤差の平均と分散を比較した結果を\figref{figure:act-control}に示す.
  なお, WTNPBを用いる際のParametric Biasは訓練時のLong/Middleの値である.
  幾何モデルが最も誤差が大きく, 関節のたわみを含むシミュレーションによる学習後, 実機による学習後の順で制御誤差が小さくなっていることがわかる.
}%

\subsection{Integrated Experiment}
\switchlanguage%
{%
  We conducted an integrated experiment that combined online update and tool-tip position control on the actual robot.
  The task involved using the tool to open a window located at a high position.
  We initiated the PB in the Short/Heavy state and the robot grasped the Long/Light tool.
  While performing random movements, the current PB was updated online and then the robot positioned itself in front of the window by walking sideways.
  The 3D position of the window was recognized from an AR marker attached to it.
  Subsequently, the robot extended the tool-tip position to the left of the window by 60 mm and then moved it 80 mm to the right to open the window.
  The entire sequence of actions was successful, demonstrating the feasibility of a series of tool manipulation tasks on a low-rigidity robot.
}%
{%
  実機においてオンライン更新と道具先端位置制御を合わせた統合実験を行う.
  道具を手に取り, 高い位置にある窓をその道具を使って開けるような動作である.
  Short/HeavyのPB値の状態から始め, Long/Lightの道具を把持する.
  ランダム動作から現在のPB値をオンライン更新し, その後横に歩いて窓の前に立つ.
  窓についたARマーカから窓の3次元位置を認識し, そこからy方向に-60 mmの地点に道具先端位置を伸ばした後, y方向に80 mm動作させ, 窓を開ける.
  一連の動作は成功し, 実機データによるWTNPBの学習, また道具状態を表すParametric Biasのオンライン更新, 道具位置と重心位置を考慮した制御により, 低剛性なロボットにおける一連の道具操作タスクが可能であることが示された.
}%

\section{Discussion} \label{sec:discussion}
\switchlanguage%
{%
  We discuss the experimental results.
  First, both the simulation and actual robot experiments have revealed that the space of PB self-organizes regularly.
  Our learning method has demonstrated the implicit embedding of tool properties such as length and weight into the network through PB.
  It is also shown that the current tool state can be updated and understood from current sensor information.
  Moreover, the accuracy of online update depends significantly on the quantity and quality of available sensors, emphasizing the importance of collecting as many sensor data as possible for higher precision and faster convergence.
  Second, control experiments in a simulation demonstrated that online update of the tool state leads to improved control precision of the tool-tip position.
  Depending on the choice of the loss function, not only the tool-tip position but also the control of the center of gravity can be achieved, allowing for whole-body tool manipulation while maintaining balance.
  Third, actual robot experiments have highlighted the significance of fine-tuning with the actual robot data.
  Control accuracy improves sequentially with training from a geometric model that does not consider deflection, incorporating deflection into the geometric model with network training, and fine-tuning with the actual robot data.
  By combining these approaches, it has been demonstrated that even low-rigidity robots can perform whole-body tool-use considering visual and tactile sensor information.

  We discuss the future prospects of this study.
  Currently, mask variables are arbitrarily determined by humans, but they should be automatically acquired by extracting correlations of sensors from experience.
  Furthermore, in robots equipped with numerous sensors, higher versatility can be achieved by automatically determining the input and output variables to be used.
  Also, while this study focused on a small humanoid robot, we anticipate that larger humanoid robots may face a variety of challenges.
  In particular, there is a need for deeper discussions on how to prevent falls and how to safely move the body randomly.
  In the future, we aim to eliminate human intervention and achieve robots with new levels of autonomy.
}%
{%
  本実験結果について考察する.
  まず, シミュレーション実験, 実機実験ともにPBの空間が規則的に自己組織化することが分かった.
  本学習方法によって, 道具の長さや重さなどのプロパティが暗黙的にネットワークに埋め込み可能であること, 現在の道具状態を多様なセンサ値から更新, 理解可能であることが分かった.
  また, そのオンライン更新の精度は, 得られるセンサの数や質によって大きく変化し, なるべく多くのセンサを集めた方が精度が高い.
  次に, シミュレーションにおける制御実験から, 道具状態のオンライン更新によって, 道具先端位置の制御精度が向上することが分かった.
  損失関数の設定次第で, 道具先端位置だけでなく, 重心位置の制御も可能であり, バランスを取りながら全身道具操作を行うことができる.
  最後に, 実機実験から, 実機データによるfine tuningが重要であることが明らかになった.
  たわみを考慮しない幾何モデル, たわみを考慮した幾何モデルによる学習, 実機データによるfine tuningの順で制御精度が向上している.
  これらを組み合わせ, 道具の状態推定と制御を組み合わせることで, 低剛性なロボットでも視覚や接触センサを考慮しつつ全身で道具操作を行うことが可能であることが示された.

  本研究の展望について述べる.
  現在は, マスク変数は人間により恣意的に決めていたが, これらは経験から相関を抜き出し, 自動的に獲得すべきものであると考えている.
  また, 多数の感覚器を備えるロボットにおいて, 用いる入出力変数さえも自動的に決定することで, より高い汎用性が得られると考えている.
  今回は小型のヒューマノイドを扱ったが, より大きなヒューマノイドでは多様な問題が起きると考えている.
  特に, 転倒をどう防ぐか, どうランダムに体を動かすかについてはより深い議論が必要である.
  今後, 人間の考える余地を排除していき, 新の自律性を持ったロボットの実現を目指していきたい.
}%

\section{CONCLUSION} \label{sec:conclusion}
\switchlanguage%
{%
  In this study, we developed a method for a low-rigidity plastic-made humanoid to learn the mutual relationship among its joint angle, center of gravity, tool-tip position, and tool-tip screen coordinates, and to consider changes in the relationship due to variations in tool weight and length for whole-body robotic tool-use.
  To facilitate the learning of the mutual relationship, we introduced a mask variable to effectively capture correlations between four different modalities, enabling the robot to adapt to partial modality data loss during online update and control.
  By utilizing Parametric Bias as a learnable input variable, the internal network representation of weight and length variations for each tool is organized without explicit labeling, allowing the robot to estimate the tool state from current sensor information.
  We updated latent variables from loss functions related to visual and tactile information, enabling precise tool-tip position control while considering deflection and center of gravity changes.
  Moving forward, we aim to explore automatic construction of mask variables and network input-output variables, as well as conduct experiments with life-sized humanoids.
}%
{%
  本研究では, 低剛性な樹脂製ヒューマノイドの道具利用について, その道具先端位置や重心, 関節角度に関する相関関係の学習を行い, さらに道具の重さや長さ変化による学習した相関関係の変化を考慮可能な手法を開発した.
  相関関係の学習にはマスク変数を導入し, 4種類のモダリティの間の相関関係を適切に捉えることで, オンライン更新や制御の際, 一部のモダリティの欠損に対応できるようになる.
  学習可能な入力変数であるParametric Biasを用いることで, 明示的なラベル付け無しに, 各道具の重さや長さの変化をネットワーク内部に自己組織化し, 現在のセンサ情報から道具の状態を推定することが可能となる.
  潜在変数を視覚や重心の情報に関する損失関数から更新し, たわみや重心変化を考慮しつつ適切に道具先端位置操作が可能となった.
  今後, マスク変数やネットワーク入出力変数の自動構築, より大きなヒューマノイドにおける実験等に取り組んでいきたい.
}%

{
  \bibliographystyle{IEEEtran}
  \bibliography{main}
}

\end{document}